\documentclass[10pt,twocolumn,letterpaper]{article}

\usepackage{cvpr}              %

\definecolor{chocolate}{rgb}{0.48, 0.25, 0.0}
\definecolor{bole}{rgb}{0.47, 0.27, 0.23}
\definecolor{goldenpoppy}{rgb}{0.99, 0.76, 0.0}
\definecolor{goldenyellow}{rgb}{1.0, 0.87, 0.0}
\definecolor{palesilver}{rgb}{0.79, 0.75, 0.73}
\definecolor{bronze}{rgb}{0.8, 0.5, 0.2}

\makeatletter
\DeclareRobustCommand\onedot{\futurelet\@let@token\@onedot}
\def\@onedot{\ifx\@let@token.\else.\null\fi\xspace}

\def\eg{\emph{e.g}\onedot} 
\def\ie{\emph{i.e}\onedot}

\makeatother

\definecolor{fuzzywuzzy}{rgb}{0.8, 0.4, 0.4}
\definecolor{MaceWindu}{rgb}{0.7, 0.1, 0.7}
\definecolor{indigo}{rgb}{0.29, 0.0, 0.51}
\definecolor{dark_green}{rgb}{0, 0.4, 0}

\def \customparskip {.3em}
\renewcommand{\paragraph}[1]{\vspace{\customparskip}\noindent\textbf{#1}}

\newcommand{\x}{\mathbf{x}}

\newcommand{\expect}{\mathbb{E}}

\DeclareMathOperator*{\argmin}{arg\,min}

\definecolor{cvprblue}{rgb}{0.21,0.49,0.74}
\usepackage[pagebackref,breaklinks,colorlinks,allcolors=cvprblue]{hyperref}

\usepackage{graphicx,subcaption} %
\usepackage{algorithm}
\usepackage{algpseudocode}
\usepackage{amsmath}
\usepackage{xcolor}
\usepackage{amssymb}
\usepackage{multirow} 
\usepackage{booktabs}
\usepackage{amssymb}
\usepackage{pifont}
\usepackage{tabularx}

\def\paperID{322} %
\def\confName{CVPR}
\def\confYear{2025}

\title{HyperNet Fields: Efficiently Training Hypernetworks without \\ Ground Truth by Learning Weight Trajectories}

\author{
  Eric Hedlin\textsuperscript{1, 2}\footnotemark[1]\quad
  Munawar Hayat\textsuperscript{2}\quad
  Fatih Porikli\textsuperscript{2}\quad
  Kwang Moo Yi\textsuperscript{1}\quad
  Shweta Mahajan\textsuperscript{2}\quad
  \\
  \textsuperscript{1} University of British Columbia\quad
  \textsuperscript{2} Qualcomm AI Research\footnotemark[2]\quad
  \\
}

\begin{document}

\twocolumn[{%
\renewcommand\twocolumn[1][]{#1}%
\maketitle
\centering
\includegraphics[width=\linewidth]{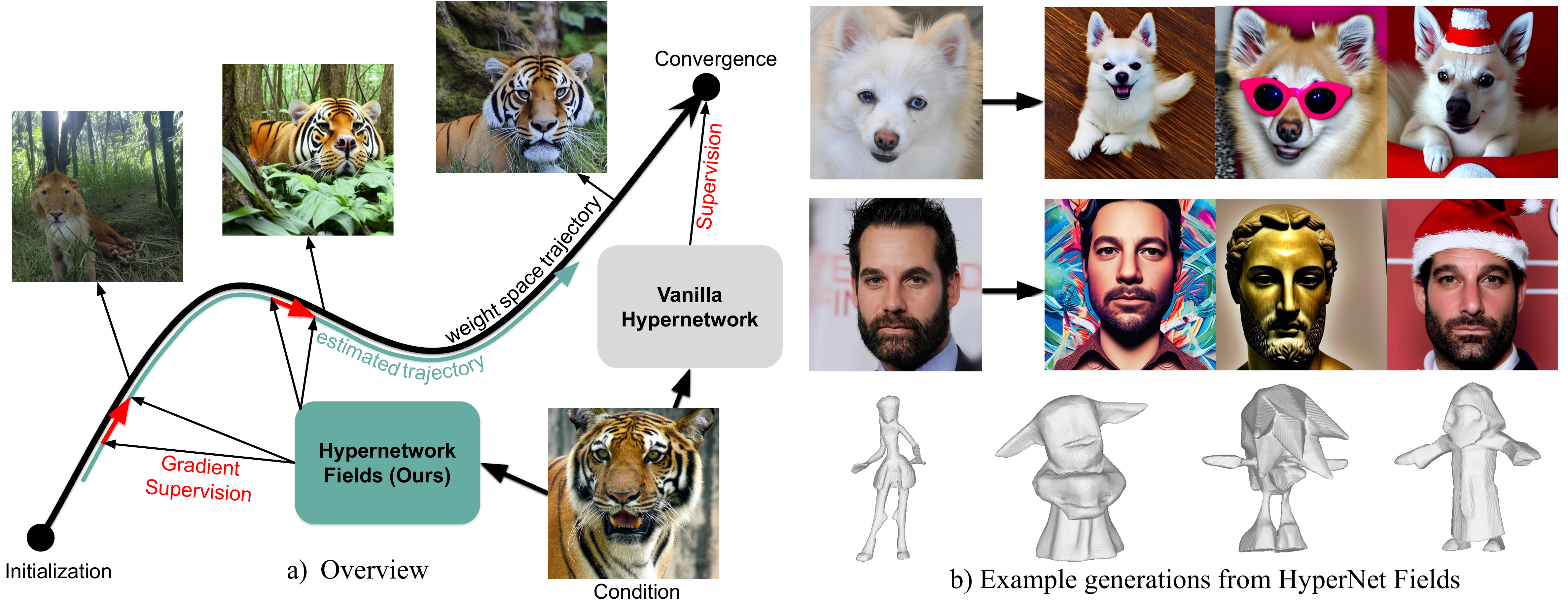}
\captionof{figure}{
    {\bf Teaser -- }
    We propose hypernetwork fields, where we learn the entire weight space trajectory instead of only the final state, which allows us to train without ever needing to know the final converged weights.
    Our method can be applied to \emph{any} application of hypernetworks, including diffusion model personalization and modeling of 3D neural representations.
    }%
    \vspace{1em}
    \label{fig:teaser}
}
]

\begin{abstract}
To efficiently adapt large models or to train generative models of neural representations, Hypernetworks have drawn interest. 
While hypernetworks work well, training them is cumbersome, and often requires ground truth optimized weights for each sample.
However, obtaining each of these weights is a training problem of its own---one needs to train, e.g., adaptation weights or even an entire neural field for hypernetworks to regress to. 
In this work, we propose a method to train hypernetworks, without the need for any per-sample ground truth.
Our key idea is to learn a Hypernetwork `Field' and estimate the entire trajectory of network weight training instead of simply its converged state.
In other words, we introduce an additional input to the Hypernetwork, the convergence state, which then makes it act as a neural field that models the entire convergence pathway of a task network.
A critical benefit in doing so is that the gradient of the estimated weights at any convergence state must then match the gradients of the original task---this constraint alone is sufficient to train the Hypernetwork Field.
We demonstrate the effectiveness of our method through the task of personalized image generation and 3D shape reconstruction from images and point clouds, demonstrating competitive results without any per-sample ground truth.
\end{abstract}
\renewcommand{\thefootnote}{\fnsymbol{footnote}}
\footnotetext[1]{Work done as part of a summer internship.}
\footnotetext[2]{Qualcomm AI Research is an initiative of Qualcomm Technologies, Inc.}

\section{Introduction}
Various generative modeling tasks~\cite{dreambooth} require adaptation, thus data-dependent network weights, or in other words, model parameters being adjusted according to various circumstances (conditions and user input).
These include personalized text-to-image generation where the weights of the underlying personalization model, such as DreamBooth~\cite{dreambooth}, are fine-tuned on the concept of interest and used to generate target images conditioned on input prompts.
Another example is 3D modeling, where neural networks are used to model shapes implicitly with signed distance fields~\cite{park2019deepsdf} or occupancy~\cite{mescheder2019occupancy}.
In both cases, a dedicated set of neural network weights is required for each personalization outcome or 3D shape.
Thus, the applicability of these methods is highly limited by the training time required to obtain these weights.

It is therefore critical to seek ways to \emph{amortize} the per-sample optimizations into Hypernetworks~\cite{hyperdiffusion,hyperdreambooth}---networks that estimate network parameters.
By `skipping ahead' the tedious training processes involved in fine-tuning or shape regressing, they greatly speed up the entire process.
Existing methods, however, have a critical shortcoming in that they supervise hypernetworks by directly matching their output to the precomputed, ground truth converged weights corresponding to the task-specific network for each sample in the dataset.
Notably, prior to any training of the Hypernetwork, one must prepare converged task-specific network weights for \emph{every} sample in the training set.
This severely limits the capabilities of Hypernetwork methods when scaling to large datasets.
As an example, to train HyperDreamBooth~\cite{hyperdreambooth}, it requires 50 days of GPU time on a NVidia RTX 3090,\footnote[3]{HyperDreamBooth~\cite{hyperdreambooth} was trained on 15k DreamBooth weights, which each take 5 minutes to compute.} to just prepare the training data alone.

Furthermore, current methods assume a one-to-one mapping between each input sample and its corresponding optimized weights. 
This assumption is an oversimplification.
Neural network optimization is stochastic with various equally plausible solutions.
Therefore, enforcing a bijective mapping between the condition and the ultimately converged weights may limit the potential expressiveness of the hypernetwork, as we will demonstrate in \cref{sec:method}.

To address these limitations, we instead propose to train a \emph{Hypernetwork Field} that estimates the entire optimization trajectory of a neural network training, rather than focusing on a single final set of weights for each sample. 
Specifically, we introduce the convergence state as additional input to the hypernetwork, such that the weights of the task-specific network at a desired state of convergence can be `queried'.
An interesting outcome of doing so is that the gradient of the estimated weights along this pathway must match the gradients of the original task-specific network, as they are both gradients pointing towards convergence.
Hence, training of such hypernetwork fields can simply be done by matching the two gradients.

By supervising the hypernetwork to match the gradient of the weights with respect to the optimization step, we avoid the need for precomputing target weights. 
This supervision strategy allows the hypernetwork to estimate a trajectory of parameters that, at each step, reflects a compatible state across all samples in the dataset.
Ultimately, at inference time, the estimated parameters which correspond with the final timestep of the hypernetwork, represent a well-converged solution for each of the samples, but are now \emph{naturally discovered} through training of the hypernetwork field itself.
Notably, this single forward pass to estimating the converged parameters is all that is necessary for inference, hence no additional compute is required compared to conventional hypernetwork approaches~\cite{hyperdreambooth,hyperdiffusion}.

Our framework is general and is applicable to \emph{all} hypernetwork training scenarios. 
To demonstrate the effectiveness of our method, we apply our method to personalized image generation and 3D shape generation.
We achieve competitive performance while reducing the training cost by $\sim$ \emph{4 times} what would be required of conventional hypernetwork training.

To summarize, our contributions are:
\begin{itemize}
    \item We propose hypernetwork fields, an approach to learn the entire optimization pathway of task-specific networks, which allows ground truth-free training without the need for precomputing the target weights;
    \item our approach drastically reduces the compute required for training hypernetworks;
    \item we validate our approach on two very different tasks and datasets---for personalized image generation we achieve results comparable to the state of the art with significantly less compute; and
    \item we further explore the application of our method to 3D shape reconstruction from images an point clouds.
\end{itemize}

\section{Related Work}
\paragraph{Hypernetworks.}
Early work on training hypernetworks backpropagates gradients through the task-specific network to the hypernetwork itself, as demonstrated by approaches \cite{ha2016hypernetworks, przewikezlikowski2024hypermaml} and subsequent explorations in continual learning \cite{oswald2019continual} and neural architecture search \cite{zhang2018graph}. 
These methods have shown that allowing gradient flow through both networks enables the hypernetwork to modify parameters for diverse tasks adaptively. 
However, more recent works have introduced approaches that involve precomputing weights for the task-specific network and directly supervise the hypernetwork to match them, eliminating the need for backpropagation through the task-specific network and improving training stability by alleviating issues with gradient flow. 
This supervision with precomputed weights has been shown to enhance gradient stability and training performance, as highlighted in frameworks like HyperDiffusion \cite{hyperdiffusion}, HyperFields \cite{babu2024hyperfields}, and HyperDreamBooth \cite{hyperdreambooth}. However, this limits the size of the dataset to which this can be applied, as the precompute requirement is proportional to the number of samples in the dataset.
Our method combines the best of both approaches by removing the need for precomputed weights while explicitly providing weight-based supervision in the form of gradients along the convergence path, enabling stable training on large datasets.

\paragraph{Denoising Models.} 
In generative modeling, finding smooth and reliable paths through latent space has become central to producing coherent samples. Examples include diffusion models \cite{sohl2015deep, ho2020denoising}, which iteratively trace a path from noise to data through successive steps, effectively navigating high-dimensional space by reversing a stochastic process. 
Consistency models \cite{cm} refine this by learning non-Markovian trajectories that reach target data distributions more efficiently, introducing a formulation allowing single-step transitions to any timestep; subsequent works \cite{kimconsistency, frans2024one} further improve this by supervising the entire denoising trajectory with guidance from a teacher model.
Flow-matching models \cite{lipman2023flow} take a similar path-based approach by learning a vector field that guides samples along pre-defined flows, ensuring a direct and smooth transition through the latent space. 
Similarly to these methods, our method can be thought of as finding a path through this high-dimensional space, specifically in the form of a single-step estimator from initialization to any desired state of convergence, whose path travels strictly along the direction of the steepest gradient descent.

\paragraph{Methods with gradient-supervision.}
Several recent approaches in generative modeling and physics-informed learning employ supervision exclusively through gradients rather than direct supervision between input-output pairs to achieve denoising and reconstruction. 
Score-based generative models \cite{song2020generative, vincent2011connection, song2021maximum} use score matching to learn the gradient (or "score") of the log density of noisy data distributions, bypassing the need for direct output supervision. 
Denoising Score Matching (DSM) \cite{vincent2011connection} trains the model to predict this gradient, guiding it to approximate data distributions through iterative denoising and demonstrating improved stability in noise removal. 
Similarly, Physics-Informed Neural Networks (PINNs) \cite{raissi2019physics, raissi2017physics1, raissi2017physics2} leverage known derivatives from partial differential equations (PDEs) in physical simulations, such as heat diffusion or fluid flow, to inform the network solely through spatial or temporal derivatives. 
This derivative-based guidance encourages convergence as the model adapts to match observed physical behaviour, requiring no explicit target outputs. 
Energy-Based Models (EBMs) \cite{grathwohl2020learning, lecun2006tutorial, du2019implicit} also align with this gradient-driven approach by learning energy landscapes that adjust gradients to represent target distributions, allowing for smooth gradient matching through learned energy fields. 
Inspired by these methods, our approach supervises only the gradient direction in task-specific weight space, effectively learning a convergence pathway without requiring explicit final state supervision. 
By supervising gradients within this pathway, our method achieves robust training stability while generalizing over diverse tasks without per-sample ground truth weights.

\section{Method}
\label{sec:method}
\begin{figure*}
\centering
\begin{minipage}{.48\textwidth}
  \centering
  \includegraphics[width=\linewidth]
  {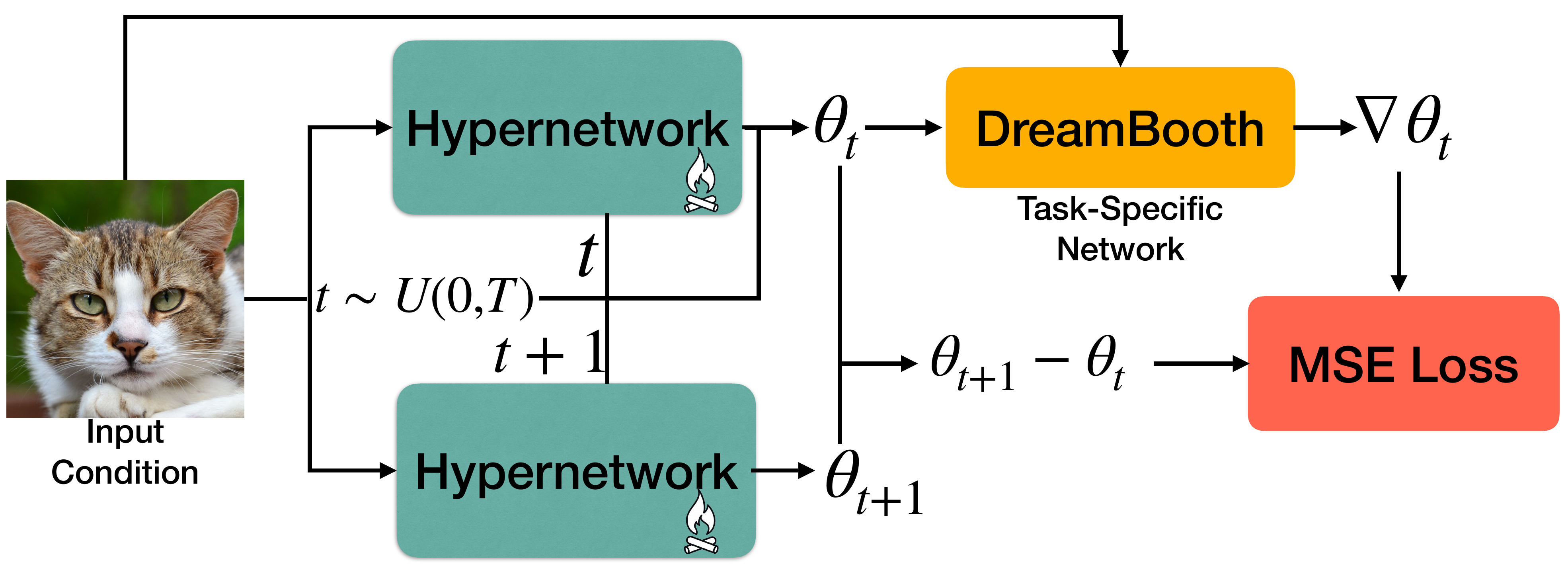}
\end{minipage}%
\hfill
\begin{minipage}{.48\textwidth}
  \centering
  \includegraphics[width=\linewidth]{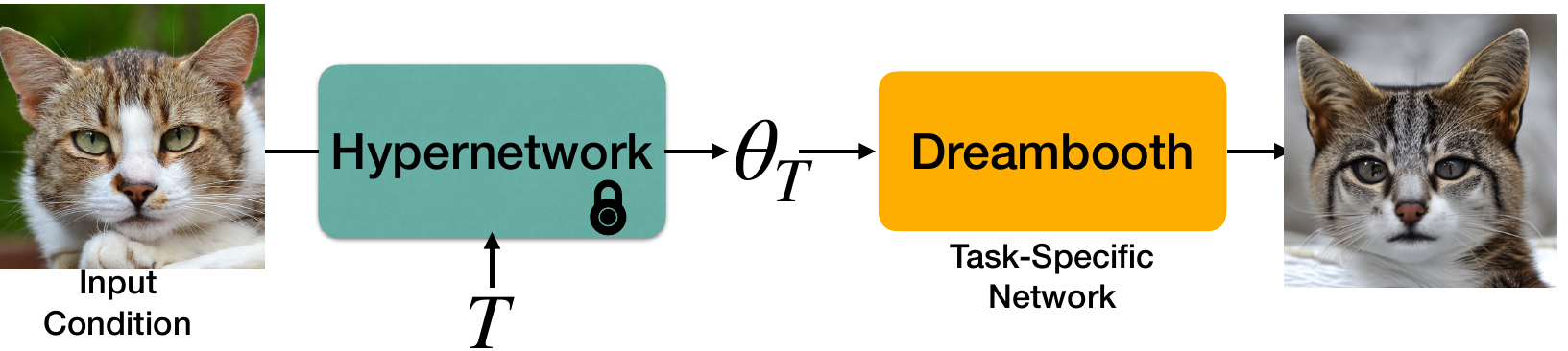}
  \label{fig:sampling}
\end{minipage}
\caption{The Hypenet Fields Framework. \emph{Left}: Our Hypernetwork with gradient-based supervision from a task-specific network -- dreambooth \cite{dreambooth}. \emph{Right}: The sampling process to generate personalized images with our hypernetwork framework.}
\label{fig:archi}
\end{figure*}

To remove the precompute requirements associated with training hypernetworks, we present a new framework that models the convergence dynamics of task-specific networks into a single, efficient process, which we name \emph{Hypernetwork Field}.

\paragraph{Problem formulation.}
Formally, let \( f_{\theta} \) denote a task-specific neural network, \eg, dreambooth \cite{dreambooth},
parameterized by \( \theta \).
$f_{\theta}$ is then typically trained to minimize a loss function \( \expect_{\mathbf{x}}\left[\mathcal{L}(\theta, \mathbf{x}) \right]\), by training for $T$ optimization steps given a dataset of conditions, \eg, image \( \mathbf{x}\). 
In our work, we wish to learn a hypernetwork field
\( H_{\phi}(t,\mathbf{x}) \), parameterized by \( \phi\) that takes as input a condition $\mathbf{x}$ and an optimization step $t$. 
The hypernetwork field then predicts the parameters \( \theta_t = H_{\phi}(t,\mathbf{x}) \) at optimization step \( t \) of the task-specific network \( f_{\theta_t}(\mathbf{x}) \), given the condition \( \mathbf{x} \) and the timestep \( t \).
In the following, we show how to obtain $H_{\phi}(t,\mathbf{x})$ without requiring any $\theta_t$ to supervise the training.%

\subsection{HyperNet Fields}

To train the hypernet field \( H_{\phi}(t,\mathbf{x}) \) without the ground-truth targets for it to regress, we take a step back and revisit what standard hypernetworks are optimizing for.
Standard approaches for learning hypernetworks \cite{hyperdiffusion,hyperdreambooth} utilize the ground-truth weights \( \theta^{*} \) of the task-specific network, corresponding to the optimized parameters for a specific sample $\mathbf{x}$ as a supervision signal. 
In other words, they first find 
\begin{equation}
    \theta^{*}(\x) = \argmin_{\theta}\mathcal{L}(\theta, \mathbf{x})
    ,
    \label{eq:original}
\end{equation}
for each $\x$, then find $\phi$, such that $\theta^{*}(\x)\approx H_\phi(\cdot, \x), \forall \x$.
This typically requires several iterations of the optimization process and thus incurs a large computational cost for generating the target parameters \( \theta^{*}(\x) \).
Moreover, since this process has to be repeated for each sample independently, the scalability to large datasets is severely limited.

Note, however, \cref{eq:original} is typically found through gradient-based optimization, and forms a trajectory of weight parameter space in the $\{\theta_t\}_{t=1\dots T}$, where $T$ is the number of optimization steps.
If we now consider our hypernet field $H_\phi(t,\x)$ that we intend to model, we notice an interesting relationship that must hold between the $\mathcal{L}(\theta_{t}, \mathbf{x})$ and $H_\phi(t,\x)$.
Specifically, consider the gradient of the task-specific weights at an optimization step $t$, \( \nabla_{\theta_{t}} \mathcal{L}(\theta_{t}, \mathbf{x}) \).
Then, if a hypernet field $H_\phi$ has been properly trained, the direction in which $\nabla_{\theta_{t}} \mathcal{L}(\theta_{t}, \mathbf{x})$ point to must match the direction in which the hypernet trajectory flows, that is, $\nabla_{t} H_\phi(t, \x)$.
In other words, their gradients must \emph{match}.

In fact, matching the two gradients allows us to circumvent the necessity of per-sample optimization as \( \nabla_{\theta_{t}} \mathcal{L}(\theta_{t}, \mathbf{x}) \) can be computed with relatively minimal overhead by using the current hypernet field estimate for $\theta_t$.
This process can also be understood as \emph{amortizing} \cref{eq:original} into the training process of the hypernet field $H_\phi(t,\x)$, as we are `skipping ahead' parts of the optimization loop of \cref{eq:original} with the estimates coming from $H_\phi(t,\x)$.
Through this amortization, %
the hypernet field captures detailed information about how the weights evolve throughout training, allowing it to effectively predict and generate weights across the training samples.
We now further detail the exact training process.

\subsection{Gradient-based Supervision}
In each training iteration of the optimization process, the hypernet field generates an estimate of the task-specific parameters \( \hat{\theta}_t = H_\phi(t, \x) \) at timestep \( t \) based on the input $\mathbf{x}$ from a dataset $\mathcal{D}$ and the timestep $t \sim U(0, T)$. 
Given this estimate, we compute the gradient of the loss function with respect to the task-specific weights at that timestep, \( \nabla_{\hat{\theta}_t} \mathcal{L}(\hat{\theta}_t, \mathbf{x}) \), and perform a single optimization step to update the weights to \( \theta_{t+1} \). 
This is given by
\begin{equation}
  \theta_{t+1} = \hat{\theta}_{t} - \eta \nabla_{\hat{\theta}_t} \mathcal{L}_{task}(\hat{\theta}_t, \mathbf{x}),
  \label{eq:gradupt}
\end{equation}
where \( \eta \) is the learning rate. 
Then, the hypernet field must also follow this update by definition, thus we wish to have $\hat{\theta}_{t+1} \approx \theta_{t+1}$.
Moreover, assuming $\theta_0 = \hat{\theta}_0 = H_{\phi}(0,\mathbf{x})$, the entire trajectory can simply be matched via minimizing 
\begin{equation}
\mathcal{L}_{\Delta}(\mathcal{D}) = \mathbb{E}_{\mathbf{x} \sim \mathcal{D}, t \sim U[0,T]}\left\| \theta_{t+1} - H_{\phi}(\mathbf{x}, t+1) \right\|^2,
\label{eq:loss_gradient}
\end{equation}
which we name as the gradient matching loss.
This loss function essentially minimizes the difference between the gradient step from \cref{eq:gradupt} and the numerical gradient $\nabla_{t}H_\phi(\mathbf{x},t)= H_\phi(\mathbf{x},t+1)-H_\phi(\mathbf{x},t)$ over the entire dataset for $T$ steps of optimization; thus replicating the underlying training dynamics of the task-specific network.

Once trained, the hypernetwork field $H_\phi(t,\x)$ thus mimics the \emph{entire} training trajectory of the task-specific network, instead of the point estimate $\theta^*(\mathbf{x})$.
We summarize this training process in \cref{fig:archi}.

\paragraph{Ensuring a consistent trajectory throughout training.}
To ensure training stability, we fix the starting position $\theta_0$ for each sample $\x$ throughout training, initializing it randomly only once at the beginning. 
This approach mirrors the typical neural network initialization to random values, followed by convergence toward the optimal parameters.

\paragraph{Forcing 0 offset at t=0 by construction.} 
Furthermore, \cref{eq:loss_gradient} matches the entire trajectory only when $\theta_0 = \hat{\theta}_0 = H_{\phi}(0,\mathbf{x})$.
Enforcing this, however, can easily be done by construction.%
Inspired by \cite{edm,cm}, for any timestep $t$, we parameterize the model with an offset from the input $\theta_0$, conditioned on $t$ as,
\begin{equation}
H_\phi(\mathbf{x}, t) = \theta_0 + \frac{t}{T} \times  H'_\phi(\mathbf{x}, t),
\end{equation}
where $H'$ is the raw hypernetwork field output from a neural network.%

\begin{algorithm}[t]
\caption{Hypernet Fields Training Algorithm}
\begin{algorithmic}[1]
\State \text{Initialize} $\theta_0$\\
$H_\phi \gets \text{Hypernetwork()}$  \Comment{Hypernetwork with parameters $\phi$}
\While{\texttt{not converged}}
    \State $\mathbf{x} \gets \text{next}(\mathcal{D})\text{}$ \Comment{Sample condition $\mathbf{x}$ from dataset}
    \State $t \gets \text{sample from } [0, T)$ 
    \State $\hat{\theta}_t \gets H_\phi(\mathbf{x}, t)$ \Comment{Estimate weights for timestep $t$}
    \State $\Delta {\theta_{t}} \gets - \eta \nabla_{\theta_t} \mathcal{L}_{task}(\hat{\theta}_t, \mathbf{x})$ \Comment{GT gradient for $\hat{\theta_{t}}$}
    \State $\Delta \hat{\theta_{t}} \gets H_\phi(\mathbf{x}, t+1) - \hat{\theta}_t$ \Comment{Est gradient for $\hat{\theta_{t}}$}
    \State $\phi \gets \phi - \eta \nabla_{\phi} \text{MSE}(\Delta {\theta_{t}}, {\Delta} \hat{\theta_{t}})$ \Comment{Update $\phi$}
\EndWhile
\end{algorithmic}
\label{algo:hypernetfield}
\end{algorithm}

\begin{figure*}
    \centering
    \includegraphics[width=\textwidth]{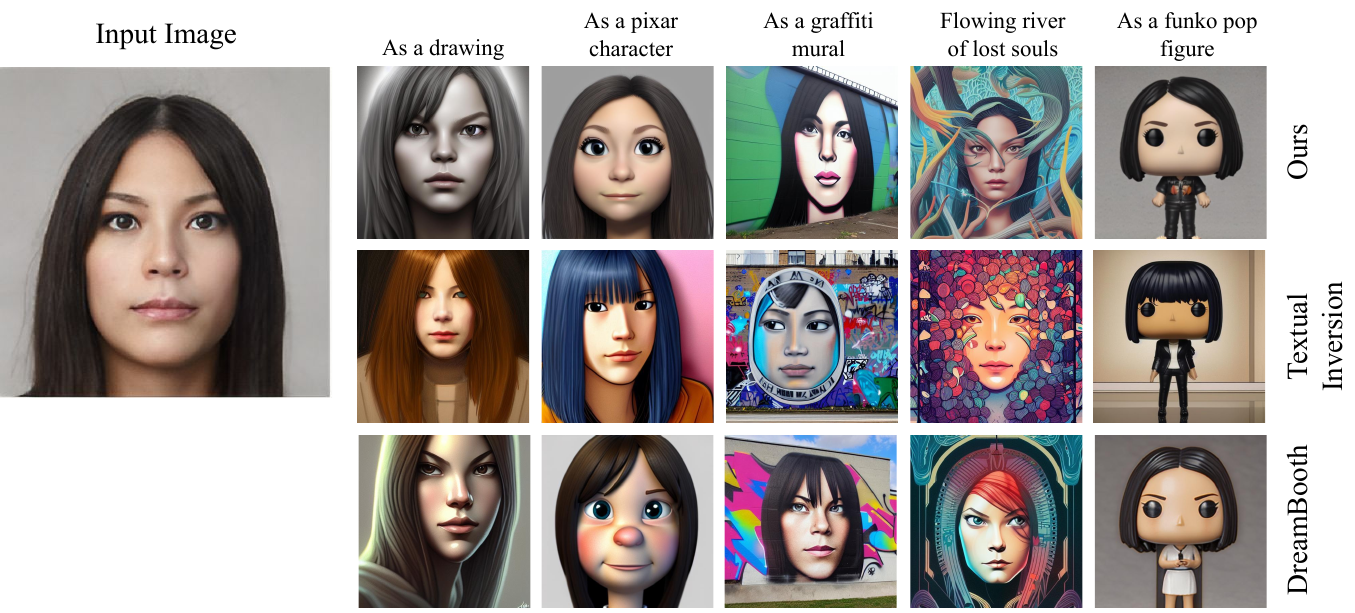}
    \caption{
    {\bf Qualitative examples (CelebA HQ) -- }
    Qualitative examples of personalized human face generation using the CelebA HQ dataset~\cite{celeba,celebahq} are shown. 
    Our hypernetwork field achieves fast adaptation, producing results comparable to DreamBooth~\cite{dreambooth} and Textual Inversion~\cite{gal2022image} while preserving individual features in a personalized manner. 
    Additional examples are provided in the supplementary material.
     }
    \label{fig:celeba}
\end{figure*}

\subsection{Algorithm}
We summarize our overall optimization process in \cref{algo:hypernetfield}. 
We first randomly initialize and fix the parameters $\theta_0$.
We then consider a random sample $\mathbf{x}$ from the dataset and a random timestep $t$ from the optimization trajectory.
With these inputs, we use the hypernetwork $H_\phi$ to obtain the weights $\hat{\theta_t}$ of the task-specific network. 
Next, using these weights and the input $\mathbf{x}$ we compute the gradient step $\Delta{\theta_t}$ for the task-specific network. 
Additionally, we calculate   $\Delta\hat{\theta_t} = \hat{\theta}_{t+1} -\hat{\theta}_t$  
from the hypernetwork.
The parameters $\phi$ of the hypernetwork are then updated to match $\Delta\hat{\theta_t}$ with the gradients $\Delta{\theta_t}$, which is equivalent to minimizing \cref{eq:loss_gradient}. 
This yields a computationally efficient framework where a single hypernetwork encodes the optimization trajectory and can be queried to get the weights of the task-specific network at any stage in the training process.

For inference, consider DreamBooth \cite{dreambooth} as a task-specific network (\cref{fig:archi}). 
To generate personalized images for a given source image, we query the hypernetwork field with the source image and the timestep $T$ to obtain the weights for the DreamBooth model.
Using these weights and the conditioning prompt, we can generate personalized images in a single step.

\subsection{Implementation Details}
\label{sec:implementation}
We use a frozen Vision Transformer (ViT) \cite{vit} to encode the images.
We then pass this encoding into a hypernetwork with a 6-layer Diffusion Transformer (DiT) \cite{dit} with adaptive layer norm conditioning \cite{adaln} for both the timestep and the image embedding.
We set the sequence length of the DiT is equal to the total number of layers in the task-specific network.
We then have a dedicated Multi-Layer Perceptron (MLP) per each layer of the task-specific network that decodes to the network weights of the target network.

\section{Experiments}

We consider two diverse applications -- image personalization and 3D shape reconstruction as target applications.
While we focus on these two, we emphasize that our hypernet fields framework does not make any assumption on the task-specific network and applies to any task.
Moreover, the definition of the loss, $L_{task}(\theta_t, c)$ is not restricted to a particular domain.
Thus, our method can be applied to these problems by modifying the task-specific optimization model.

\subsection{Application to Personalized Image Generation}

To apply our method to personalized image generation, we follow prior work~\cite{hyperdreambooth} 
and evaluate and compare personalized face generation on CelebA HQ \cite{celeba, celebahq}, as well as on animal faces with the AFHQ dataset~\cite{afhq}.

\begin{figure*}
    \centering
    \includegraphics[width=\linewidth]{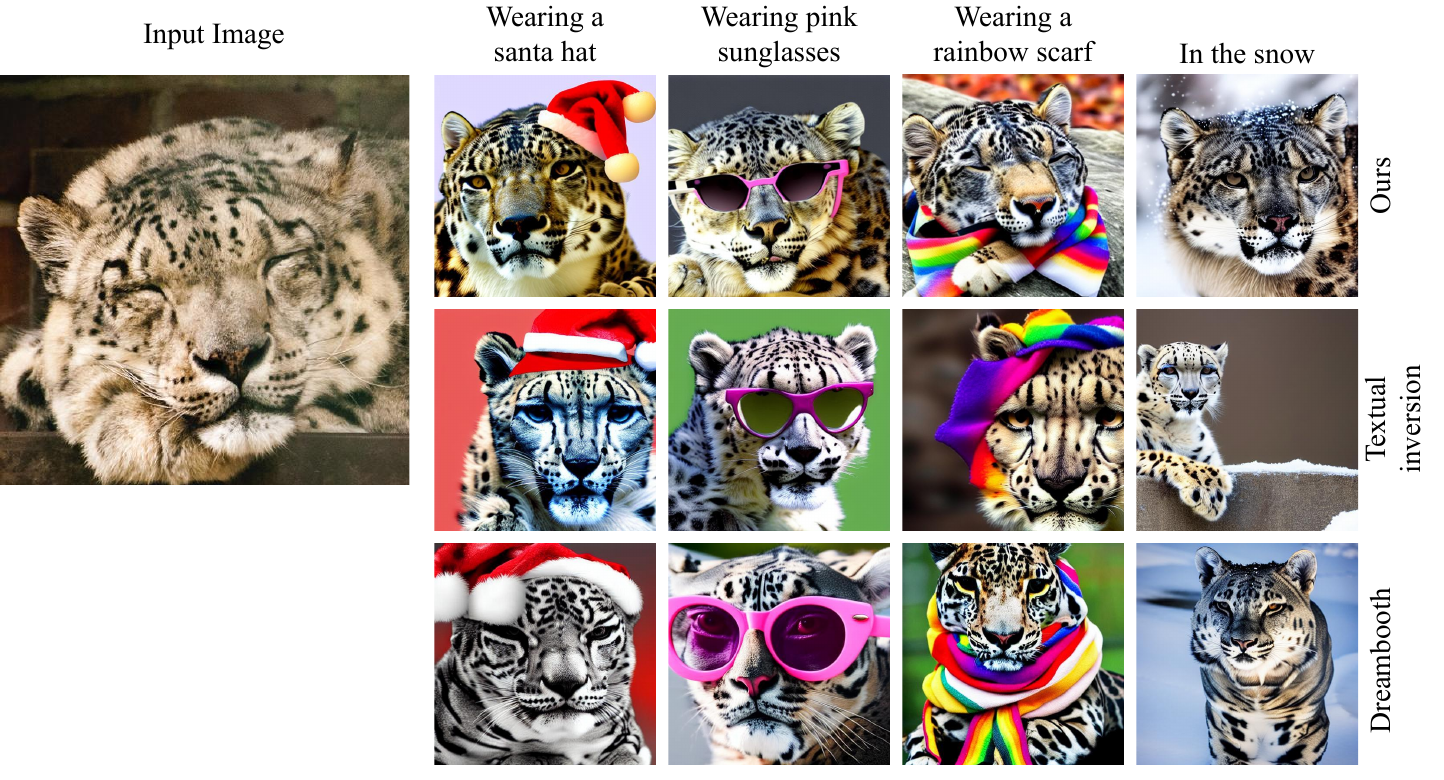}
    \caption{
    {\bf Qualitative examples (AHFQ) -- }
    Qualitative results of personalized animal image generation from the AFHQ dataset~\cite{afhq} demonstrate the adaptability of our hypernetwork field. Compared to DreamBooth~\cite{dreambooth} and Textual Inversion~\cite{gal2022image}, our method effectively captures the specific animal's characteristics while enabling rapid adaptation. Further examples are available in the supplementary material.
     }
    \label{fig:afhq}
\end{figure*}

\paragraph{Task network -- DreamBooth.} 
To personalize and generate customized images we use DreamBooth as our task network, as in HyperDreamBooth~\cite{hyperdreambooth}.
DreamBooth
builds upon the denoising diffusion probabilistic models~\cite{ldm,imagen}.
The generation process is guided by a special conditioning token encoding the characteristics of the input conditioning image, which is found by optimizing through diffusion models, which require a non-trivial amount of compute.
On the CelebA dataset~\cite{celeba} we follow prior work~\cite{hyperdreambooth} and use the text condition ``a \texttt{[V]} face"  where the token \texttt{[V]} represents a placeholder for the specific identity being learned. 
Analogously, for the AFHQ dataset \cite{afhq} with animal images, we use the text condition ``a \texttt{[V]} animal" to guide the model. 
The conditioning token along with the input prompts are then used to synthesize personalized images.

\paragraph{Training setup.}
For the CelebA-HQ dataset~\cite{celeba, celebahq}, we follow prior work~\cite{hyperdreambooth} and replicate their setup to make results comparable.
Specifically, among the 30k high-quality aligned faces, we use the train/test splits and evaluation prompts from HyperDreamBooth~\cite{hyperdreambooth}, with 29.9k images in the training set and 100 images in the test set.
For the AFHQ dataset~\cite{afhq} we use the live subject prompts from the DreamBooth dataset~\cite{dreambooth} so that the results are comparable.
Among the 16.1k aligned animal faces, we use the established train and test splits, leaving 14.6k images for training and 1.5k for testing.
In addition to the DreamBooth and HyperDreamBooth approach, we compare against Textual Inversion~\cite{gal2022image}.

For both experiments, we set the number of optimization steps to be $T=500$.
Our network is trained to output rank-1 Low-Rank Adaptation (LoRA) \cite{lora} weights for both the UNet and the text encoder for Stable Diffusion 1.5 \cite{ldm}, which amounts to 273k parameters across 352 layers.
We train our model with a batch size of 256 and for 20k iterations on 8 NVidia A100 GPUs, which takes 1.5 days.
Similar to the HyperDreamBooth~\cite{hyperdreambooth} approach once our model is trained, we can perform 50 steps of fast fine-tuning starting from the hypernetwork outputs.

Note that, unlike existing works that require per-instance fitting, with our method, it is possible to introduce on-the-fly data augmentation.
We thus incorporate color, saturation, contrast, and hue augmentations to perturb the appearance of the conditioning signal $\x$, including horizontal flips.

\paragraph{Evaluation metrics.}
We use the standard metrics suggested by DreamBooth~\cite{hyperdreambooth}: CLIP-T, CLIP-I~\cite{clip}, and DINO~\cite{dino}.
CLIP-T measures the alignment between the generated image and the textual prompt, while CLIP-I and DINO evaluate the visual consistency of the generated image with the reference object’s appearance. 
We also include the identity preservation metric (Face Rec.) suggested by HyperDreamBooth~\cite{hyperdreambooth}.

\begin{table}
\centering
\resizebox{\linewidth}{!}{
\setlength{\tabcolsep}{4pt}
\begin{tabular}{@{}l cccc@{}}
\toprule
                    Method &  Face Rec.$\uparrow$ &  DINO$\uparrow$ &  CLIP-I$\uparrow$ &  CLIP-T$\uparrow$ \\
\midrule
           Hyper-DreamBooth & \textbf{0.655} &   0.473 &     0.577 &     \textbf{0.286} \\
                DreamBooth & 0.618 &   0.441 &     0.546 &     0.282 \\
         Textual Inversion & 0.623 &   0.289 &     0.472 &     0.277 \\
        Ours & 0.325 &   \textbf{0.605} &     \textbf{0.639} &     0.268 \\
\bottomrule
\end{tabular}
}
\caption{
    {\bf Quantitative results (CelebA HQ) -- }
    We report the standard metrics for our method and the baselines.
    Our method is able to offer comparable performance, with significantly less compute.
    In terms of the image alignment metrics, our method significantly outperforms other methods.
}
\label{tab:celeba}
\end{table}

\begin{table}
\centering
\begin{tabularx}{\linewidth}{@{}X ccc@{}}
\toprule
\textbf{Method} & DINO$\uparrow$ & CLIP-I$\uparrow$ & CLIP-T$\uparrow$ \\
\midrule
Textual Inversion & 0.596 & 0.780 & \textbf{0.285} \\
DreamBooth       & 0.560 & 0.763 & 0.268 \\
Ours & \textbf{0.664} & \textbf{0.807} & 0.277 \\
\bottomrule
\end{tabularx}
\caption{
    {\bf Quantitative results (AFHQ) -- }
    We outperform other baselines in terms of image alignment and provide comparable results in terms of CLIP-T, the prompt alignment.
    We note that our method outperforms DreamBooth~\cite{dreambooth} for all metrics, benefiting from the the fact that our HyperNet Field learns a general trend of how DreamBooth weights behave.
}
\label{tab:afhq}
\end{table}

\paragraph{Compute time.} As noted earlier, our method requires 1.5 days to train with 8 GPUs for the AFHQ and CelebA datasets, with around 14.6k samples for AFHQ and 29.9k samples for CelebA, amounting to 12 days of GPU time. 
This is much faster compared to the training of HyperDreamBooth~\cite{hyperdreambooth} that requires 15k DreamBooth weights to be precomputed, which amounts to 50 days of GPU time, before their hypernetwork can start training. 
Thus, our method trains in less than a quarter of the pre-processing time of HyperDreamBooth.

Since we query our Hypernetwork at a single optimization step $T$, our method also yields efficient inference.
Inference of our method takes only 0.3 seconds, compared to 2 seconds for HyperDreamBooth~\cite{hyperdreambooth} and 5 minutes for DreamBooth. 
Additionally, the fast fine-tuning required by both our method and HyperDreamBooth takes just 20 seconds, allowing for quick adaptation with minimal overhead.

\begin{figure}
    \centering
    \includegraphics[width=0.45\textwidth]{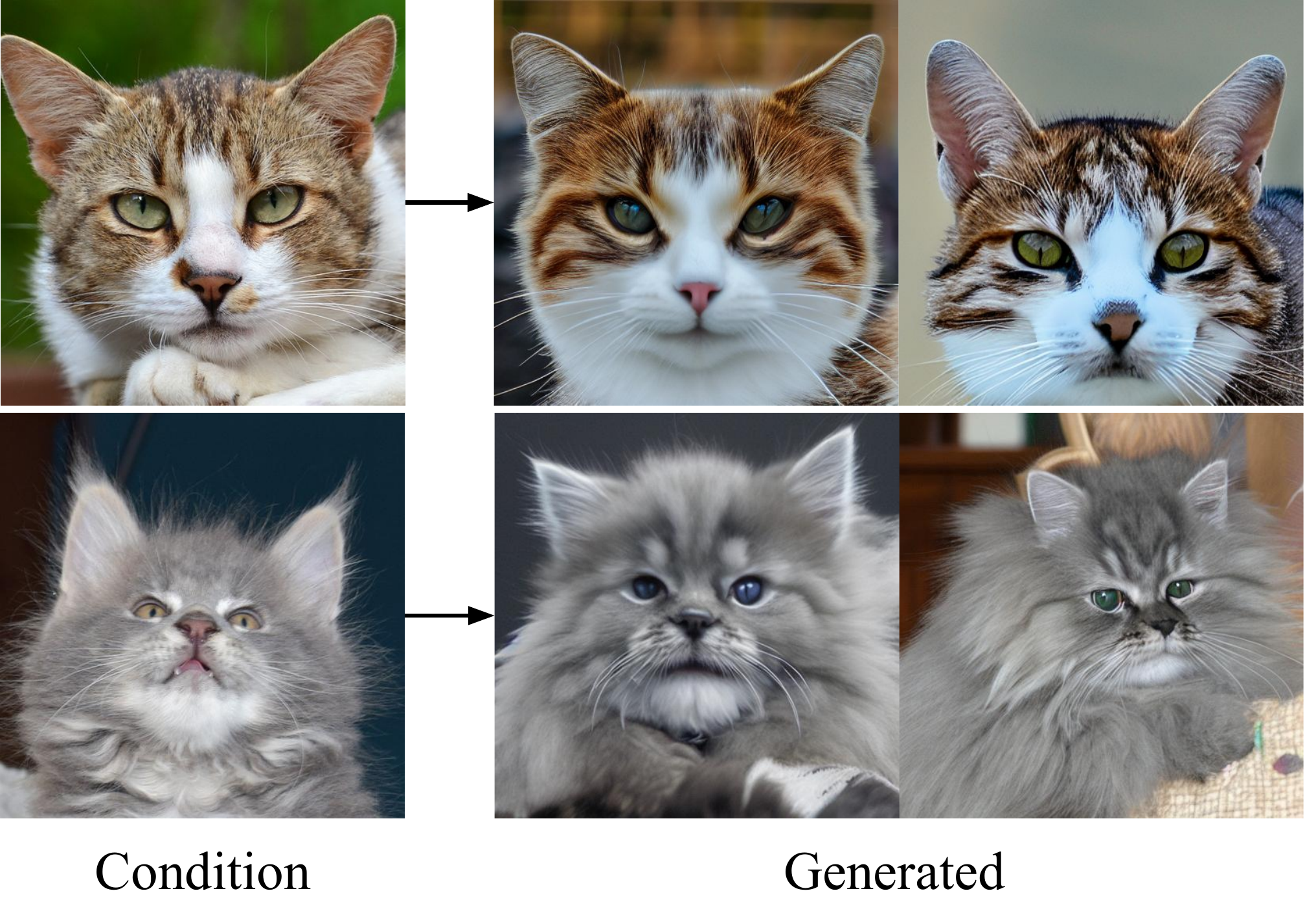}

    \caption{
        {\bf Fast training (AFHQ) --}
        We show example outcomes of our method only after 100 iterations of training, as raw output from our Hypernetwork without fast fine-tuning.
        As shown, even after only 100 iterations, our model is able to generate images that are visually very similar to the conditioning images.
    }
    
    \label{fig:fast_convergence}
\end{figure}

\paragraph{Results.} 
We provide qualitative highlights in \cref{fig:celeba,fig:afhq}, and the quantitative summaries in Tables~\ref{tab:celeba}~and~\ref{tab:afhq}.
Our results for image personalization in \cref{fig:celeba,fig:afhq} is on par with the target Dreambooth model whose weights we estimate with our HyperNet Fields approach. 
That is, our approach generates personalized images that are representative of the given input prompt and the conditional image.
Again, note that we achieve these personalization results in a single step by querying the hypernetwork once at timestep $T$.
Interestingly, our method demonstrates remarkable performance for preserving the original image context, as measured by DINO and CLIP-I scores (\cref{tab:celeba}).
Further,  in the case of the AFHQ dataset (\cref{tab:afhq}), we outperform all DreamBooth variants, demonstrating that not only do we remove the compute-heavy optimization requirement of DreamBooth~\cite{dreambooth}, but also by amortizing the behavior of the learned weights with respect to the input conditioning image, we successfully replicate the dynamics of the task-specific dreambooth network yielding improved results.
We emphasize that while results are comparable to each other, our method is significantly faster than the baselines.

\begin{figure}
    \includegraphics[width=\linewidth]{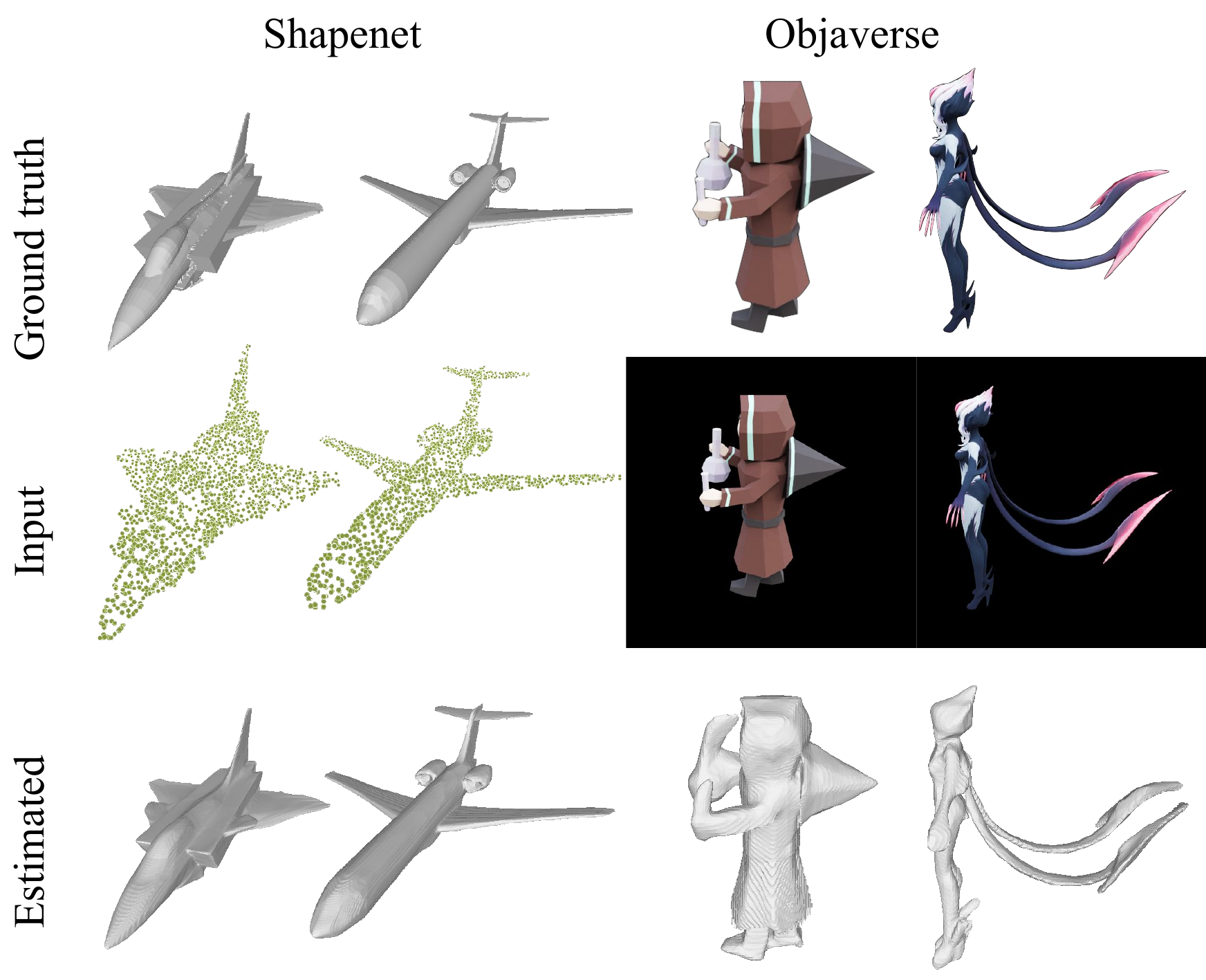}
    \caption{
        {\bf Qualitative examples (3D Shapes) --}
        We show example shape reconstruction results converting (left) point clouds to 3D shapes and (right) images to 3D shapes.
        More qualitative examples are available in the supplementary material.
    }
    \label{fig:shapes}
\end{figure}

\begin{figure*}
    \centering
    \begin{subfigure}[b]{0.08\textwidth}
        \centering
        \includegraphics[width=\textwidth]{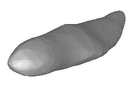} %
        \caption{$t=0$}
        \label{fig:image1}
    \end{subfigure}
    \hfill
    \begin{subfigure}[b]{0.20\textwidth}
        \centering
        \includegraphics[width=\textwidth]{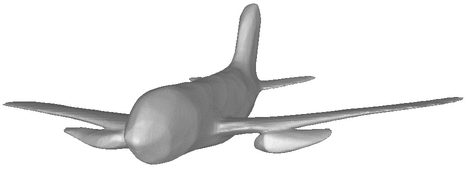} %
        \caption{$t=250$}
        \label{fig:image2}
    \end{subfigure}
    \hfill
    \begin{subfigure}[b]{0.20\textwidth}
        \centering
        \includegraphics[width=\textwidth]{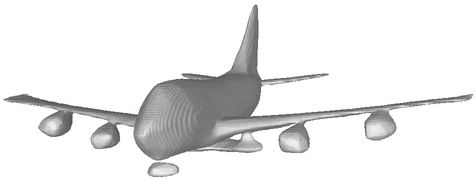} %
        \caption{$t=500$}
        \label{fig:image2}
    \end{subfigure}
    \hfill
    \begin{subfigure}[b]{0.20\textwidth}
        \centering
        \includegraphics[width=\textwidth]{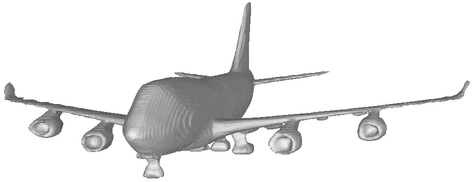} %
        \caption{$t=1000$}
        \label{fig:image2}
    \end{subfigure}
    \hfill
    \begin{subfigure}[b]{0.28\textwidth}
        \centering
        \includegraphics[width=\textwidth]{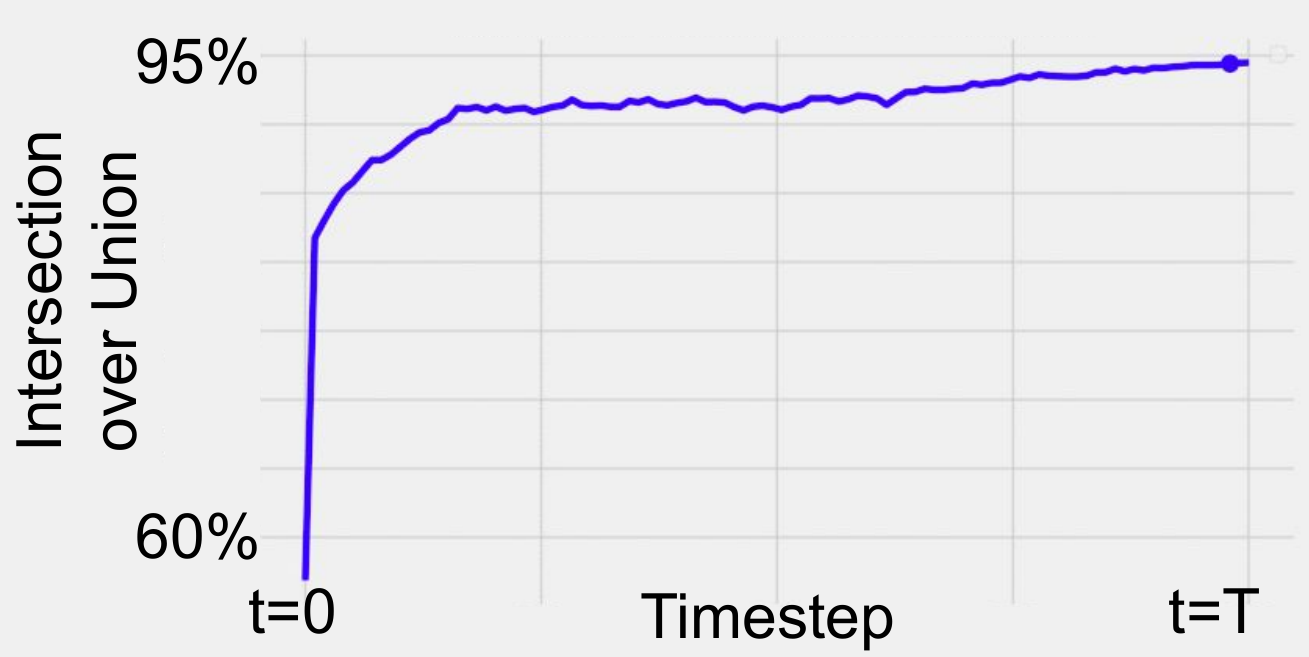} %
        \caption{IOUs over sampled Ts}
        \label{fig:image3}
    \end{subfigure}
    \hfill
    \caption{
        {\bf Hypernet Field output example (ShapeNet) --}
            We show the output of our HyperNet Field for various points along the optimization trajectory as well as the intersection over union (IoU) over all values of T.
            As shown, the HyperNet Field is able to accurately estimate the entire optimization trajectory, starting from network weights that only represent the airplane shape roughly, and converging to one that provides a detailed reconstruction of the airplane.
            The IoU plot also shows that as we traverse the trajectory, the IoU of the generated shapes to the ground truth shape improves, demonstrating convergence.
            Despite being shown only one ground truth optimization step at a time, our model is able to represent the entire optimization trajectory from initialization to convergence.
    }
    \label{fig:shapnet_t}
\end{figure*}

\paragraph{Fast training.} An interesting observation from our training process is that our model exhibits faster adaptation to personalization. 
As shown in \cref{fig:fast_convergence}, after only 100 training iterations, our model generates images that are visually similar to the conditioning images, achieving this level of personalization with minimal training time. We hypothesize that this rapid adaptation is due to our model’s ability to simulate the optimization pathway, effectively learning the direction in which to adjust network weights to reduce the task-specific loss.

\subsection{Application to 3D shape reconstruction}

We further explore the potential of our method for the task of 3D shape reconstruction, specifically the task of estimating the weights of an occupancy network~\cite{mescheder2019occupancy}.
Additionally, we consider two different conditioning inputs: a rendered image of the shape \cite{liu2023zero, jun2023shap}, and a point cloud of the shape \cite{achlioptas2018learning, qi2017pointnet, groueix2018atlasnet}.
For the image conditioning, we use the 38 renderings provided in \citet{objaverseRendering}---covering various angles around each object---which we sample one of and encode with a pre-trained ViT~\cite{vit}.
For the point cloud conditioning, we randomly sample 2048 points uniformly from the surface of the ground truth mesh and encode the input point cloud with an off-the-shelf variational autoencoder~\cite{pointflow}.

\paragraph{Task network -- Occupancy Net.}
For both cases, the hypernetwork field architecture remains the same as described in \cref{sec:implementation}, and for the task-specific network, we follow an architecture similar to HyperDiffusion~\cite{hyperdiffusion} ---we use a simple 3-layer dense network~\cite{theGoat} with ReLU activations and a hidden dimension of 128.
The task-specific network, the occupancy network, takes the 3D coordinates of a query point as input and outputs a scalar value representing the occupancy as in \citet{mescheder2019occupancy}.
We use marching cubes~\cite{marchingCubes} to render the outcome.

\paragraph{Training setup.}
To supervise the training, we 
sample 20k query points, 10k uniformly, and 10k near the ground-truth surface to be used for training.
To find the ground-truth occupancy values for these points, we first make the meshes watertight with \cite{manifoldplus} and use winding numbers \cite{windingNumbers} on the watertight mesh to determine whether each point lies inside or outside the surface, and compute the occupancy via \citet{windingNumbers}.
We use the mean squared error (MSE) of the occupancy estimates for the task-specific loss.
For all cases, we set $T=10,000$.

\paragraph{2D image conditioning (Objaverse shapes).} 
For 2D image conditioning, as training over the entire Objaverse dataset \cite{objaverse} would be computationally expensive, we simply overfit to a subset of 128 shapes, as we are mostly interested in the potential of our method.
We leave training on the full Objaverse dataset as future work. 
We obtain reasonable shape reconstruction results as shown in Figures~\ref{fig:teaser}~and~\ref{fig:shapes} with our method for this simple overfitting demonstration just within 10 hours of training on a single NVidia RTX 3090 GPU.

\paragraph{3D point cloud conditioning (ShapeNet shapes).}
For 3D point cloud conditioning, we use the entire dataset.
As shown in \cref{fig:shapes}, our method is able to learn to generate occupancy network weights from point clouds.
We further illustrate how both our Hypernet Field output and the IoU for shape generation evolves over the optimization process in \cref{fig:shapnet_t}.
As shown, our method simulates the entire training process of the task-specific network for a given point cloud despite only being shown a single ground truth gradient step at a time.

\section{Conclusion}
\label{sec:conclusion}

We presented Hypernetwork Fields, a novel and efficient hypernetwork training framework. 
Instead of optimizing individual weights per sample, we use gradient supervision and match gradients along the hypernetworks field and the task-specific network at random points along the optimization trajectory.
This drastically reduces the precomputation overhead of training hypernetworks, while maintaining comparable performance to state of the art.
We have demonstrated the effectiveness of our method on personalized image generation and 3D shape reconstruction.
Our framework is general and can be applied to any hypernetwork training scenario, making it a promising direction for scaling hypernetwork applications to larger datasets.

\paragraph{Limitations and future work.}
Our method enables stable hypernetwork training on large datasets.
Because of this, the potential of our method for diverse tasks remains underexplored.
This may include tasks such as training a hypernetwork to estimate the weights of NeRFs or even personalized adapters for large language models.
However, our approach requires as many output neurons as there are weights in the target network, resulting in suboptimal memory usage. Recent work by \citet{dravid2024interpreting} has shown that personalized DreamBooth adapters can be represented using only 1k parameters. 
Integrating our method with this compact representation approach could significantly enhance parameter efficiency while maintaining our proposed benefits of efficient hypernetwork training.
These would be interesting avenues for future work.

{
    \small
    \bibliographystyle{ieeenat_fullname}
    \bibliography{macros.bib, main}
}

\clearpage

\section{Ablations}

To further motivate our design choices, we show additional ablation results.

\begin{figure}[htbp]
    \centering
    \begin{subfigure}[b]{0.23\textwidth}
        \includegraphics[width=\textwidth]{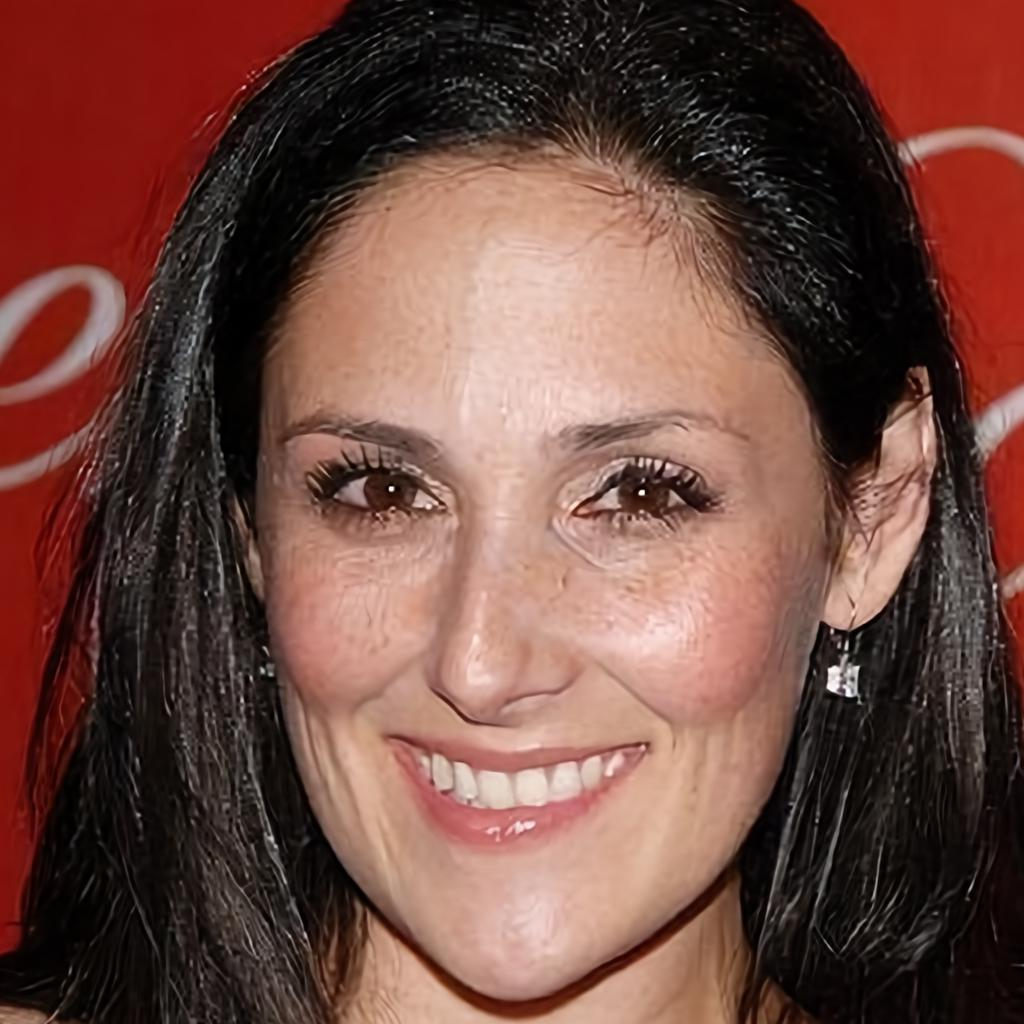}
        \caption{Condition image}
        \label{fig:image1}
    \end{subfigure}
    \hfill
    \begin{subfigure}[b]{0.23\textwidth}
        \includegraphics[width=\textwidth]{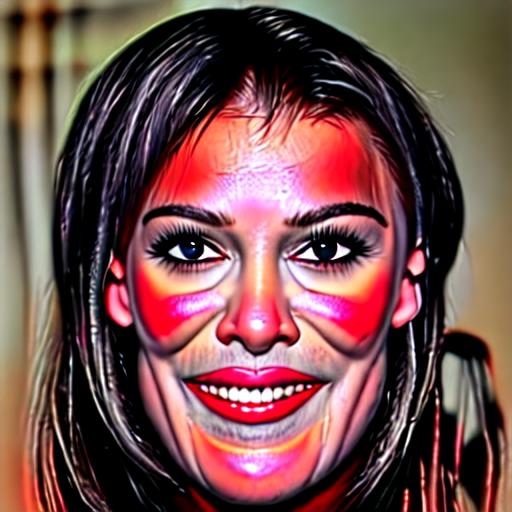}
        \caption{"a [V] face"}
        \label{fig:image2}
    \end{subfigure}
    \hfill
    \begin{subfigure}[b]{0.23\textwidth}
        \includegraphics[width=\textwidth]{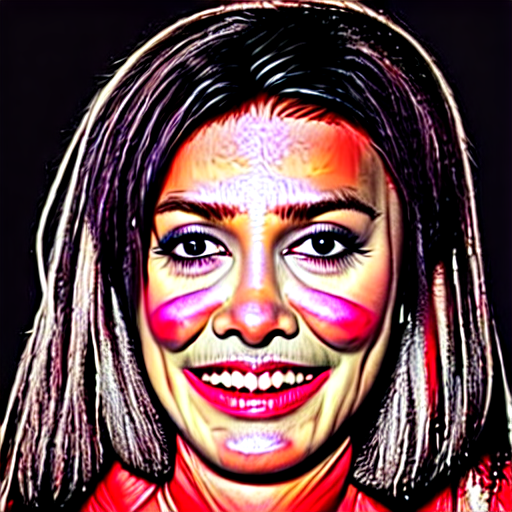}
        \caption{"... wearing a santa hat"}
        \label{fig:image3}
    \end{subfigure}
    \hfill
    \begin{subfigure}[b]{0.23\textwidth}
        \includegraphics[width=\textwidth]{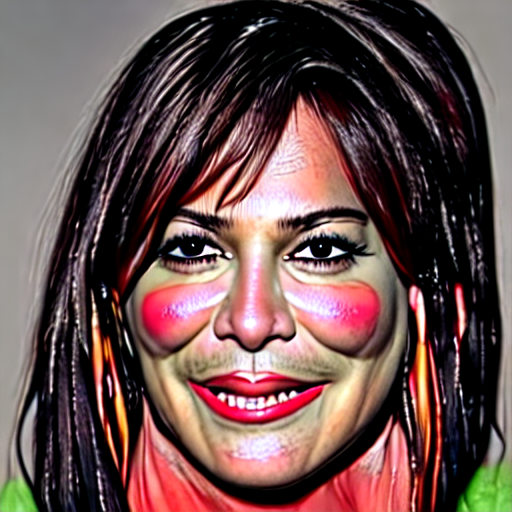}
        \caption{"... as a pixar character"}
        \label{fig:image4}
    \end{subfigure}
    \caption{
    \textbf{Example results of training with the task-specific loss --}
    Training directly with the task-specific loss results in the hypernetworks training simply learning to overfit to a specific training sample, and ignoring the user prompt.
    As shown, results start being irrelevant to the prompt.
    Also this further results in unstable training, and thus the results shown with red artifacts on faces.
    }
    \label{fig:recon_ablation}
\end{figure}

\paragraph{Reconstruction loss.}
Specifically, we further compare using our method against using an identical setup as ours, but directly learning the hypernetwork through the task-specific loss.
More formally this can be written as updating parameters $\phi$ in the following equation:
\begin{equation}
    \phi \gets \phi - \eta \nabla_{\theta_{H_\phi(c, T)}} L_{task}(H_\phi(c, T), c).
\end{equation}
While this looks like a good idea at first glance training results in over-adherence to the single example that was given---it results in outputting the same image regardless of the prompt, as shown in \cref{fig:recon_ablation}.
We report the quantitative performance of optimizing using the task-specific loss, \ie, the reconstruction loss in \cref{tab:celeba_ablations} for the CelebA dataset and in \cref{tab:afhq_ablations} for the AFHQ dataset.
Note how the CLIP-T scores are much worse than with our method, which effectively shows what happens in \cref{fig:recon_ablation}.

\paragraph{Fast fine tuning (FFT).}
We further report our performance without the fast fine-tuning applied to our model.
As shown, there is a slight decrease in performance in terms of CLIP-I and DINO, but both are still much higher in the case of CelebA than the baselines, and for AFHQ, still comparable.
Note, however, how CLIP-T scores are higher, compared to any of the baselines, and even our method with FFT.
This actually indicates better adherence to prompts, as demonstrated earlier in \cref{fig:celeba} and in \cref{fig:afhq}.
We further argue that low Face Recognition score in \cref{tab:celeba_ablations} is not a critical factor, as once they are transferred to different contexts, such as the funko pop figure in \cref{fig:celeba}, it is natural that they are not high.

\section{Extra Qualitative Results}
\begin{figure}
    \centering
    \includegraphics[height=0.95\textheight]{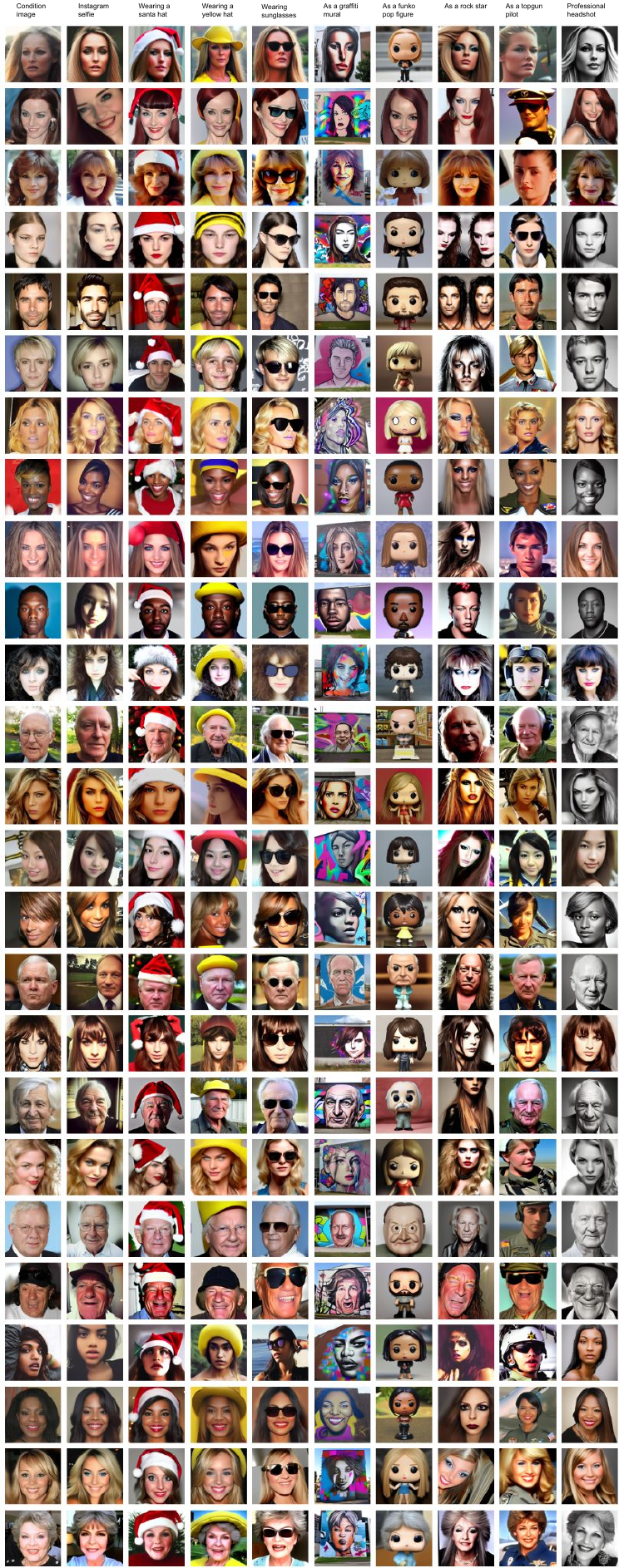}
    \caption{
    {\bf Celeba without fast fine tuning -- }
    Each of these images had their dreambooth parameters estimated in a single forward pass of our hypernetwork field. Our model can be seen to produce reasonable results even without extra fine tuning.
     }
    \label{fig:celeba_no_fft}
\end{figure}

\begin{figure}
    \centering
    \includegraphics[height=0.95\textheight]{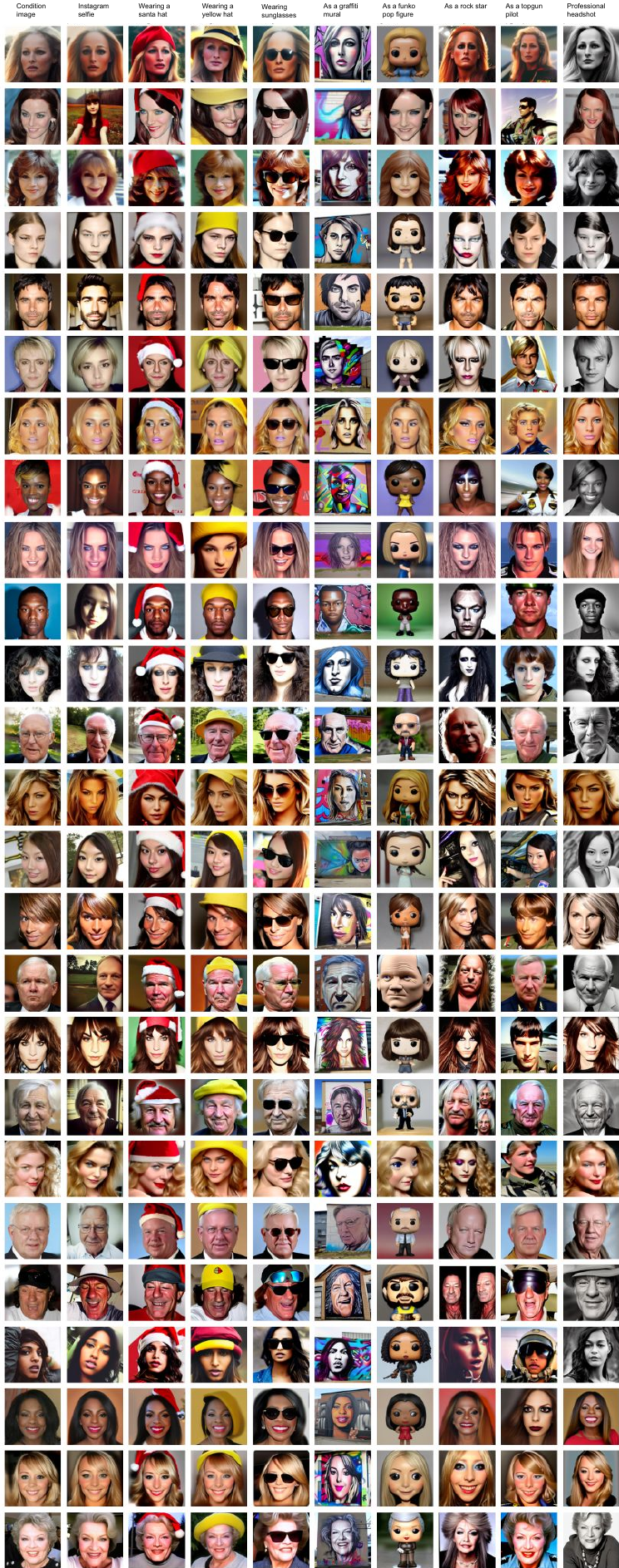}
    \caption{
    {\bf Celeba with fast fine tuning -- }
    Each of these images had 50 iterations of DreamBooth fast fine tuning applied to them.
     }
    \label{fig:celeba_w_fft}
\end{figure}

\begin{figure}
    \centering
    \includegraphics[height=0.95\textheight]{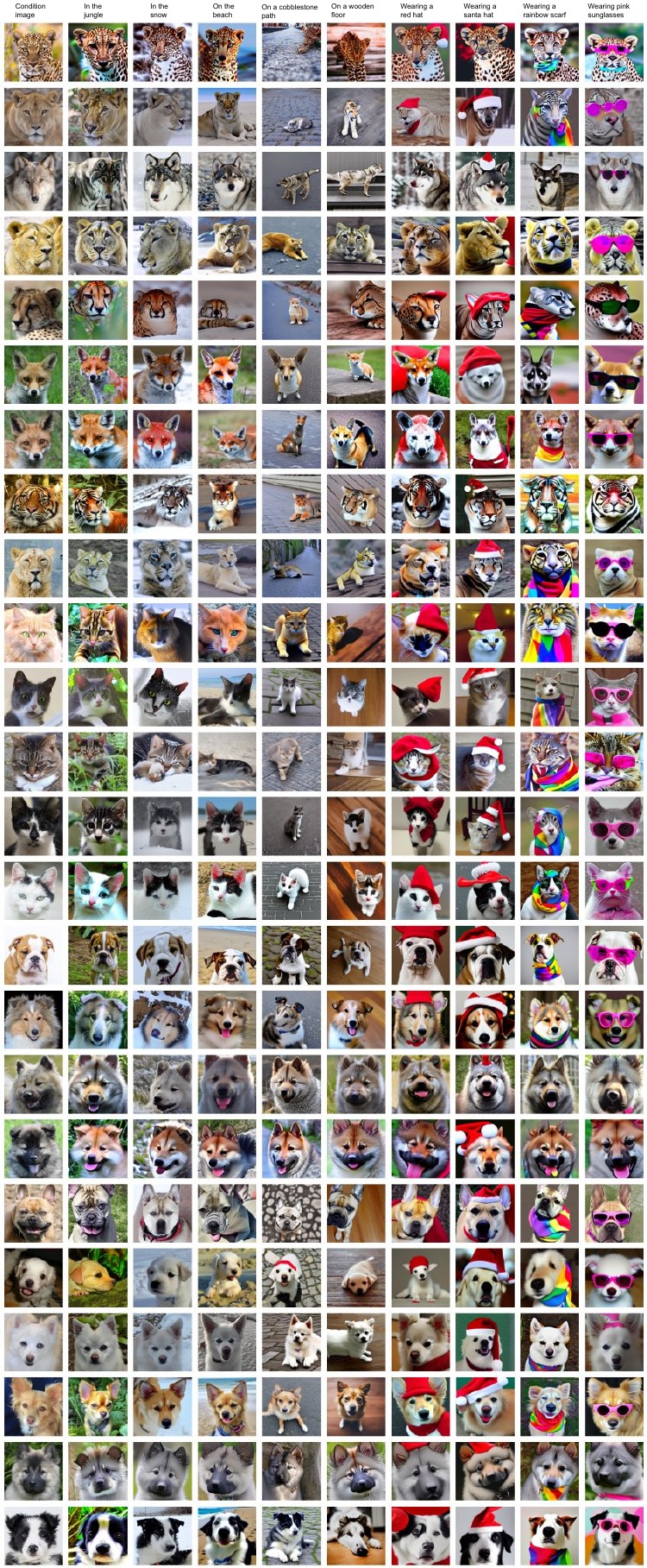}
    \caption{
    {\bf AFHQ without fast fine tuning -- }
    Each of these images had their dreambooth parameters estimated in a single forward pass of our hypernetwork field. Our model can be seen to produce reasonable results even without extra fine tuning.
     }
    \label{fig:afhq_no_fft}
\end{figure}

\begin{figure}
    \centering
    \includegraphics[height=0.95\textheight]{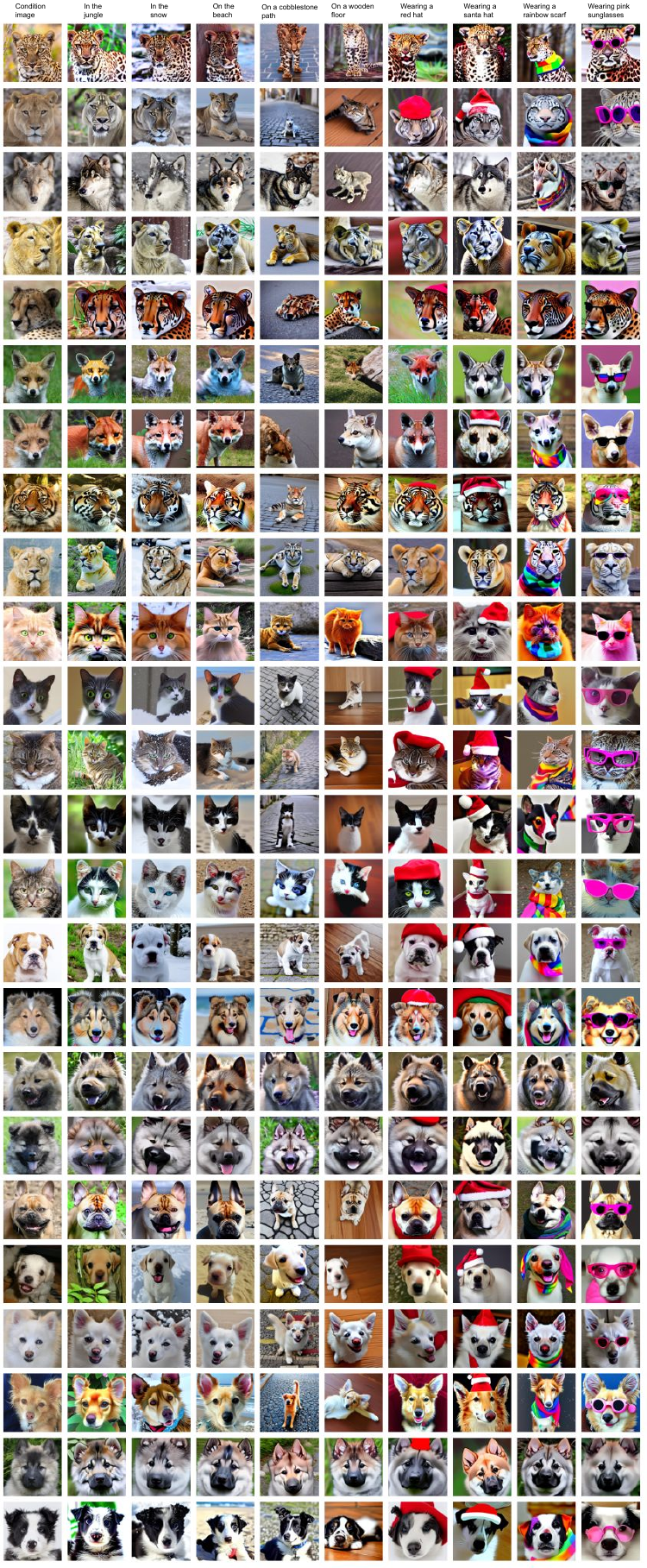}
    \caption{
    {\bf AFHQ with fast fine tuning -- }
    Each of these images had 50 iterations of DreamBooth fast fine tuning applied to them.
     }
    \label{fig:afhq_w_fft}
\end{figure}

We present additional results generated by our hypernetwork, both in its direct output form and after fast fine-tuning. Specifically, we include images from the AFHQ dataset \cite{afhq} sampled directly from the hypernetwork (Figure ~\ref{fig:afhq_no_fft}) and after fast fine-tuning (Figure ~\ref{fig:afhq_w_fft}). Similarly, we show results for the CelebA dataset \cite{celeba} directly from the hypernetwork (Figure ~\ref{fig:celeba_no_fft}) and after fast fine-tuning (Figure ~\ref{fig:celeba_w_fft}).
These results demonstrate that our hypernetwork produces reasonable outputs without additional tuning, but higher-quality images can be achieved with the fast fine-tuning process.

\paragraph{User Study.} Automatic face reconstruction metrics have inherent limitations, as they are primarily designed for \emph{aligned, photorealistic human faces}. Consequently, their reliability diminishes significantly when evaluating stylized or abstract prompts, such as 'as funko pop figure' or 'as a graffiti mural.' To address this limitation, we supplement the automatic metrics with a human preference survey, providing a more nuanced assessment of how well each method preserves the identity of the subject across diverse prompts.

In this user study, participants were asked to indicate their preferred image generation method, given the conditioning image and a specific prompt. Responses were collected from a total of 903 evaluations. The summarized preference scores are presented in \cref{fig:user_study}, clearly indicating that our method is significantly favored compared to baseline approaches.

While automatic metrics are included for completeness, these human preference results offer a more reliable indicator of method effectiveness, particularly highlighting the robustness of our method in maintaining subject identity across stylized or abstract image generation scenarios.

\begin{table}[h!]
\centering
\begin{tabularx}{\linewidth}
{@{}Xcccc@{}}
\toprule
\textbf{Ablation} & Face Rec. & CLIP-I & DINO & CLIP-T \\
\midrule
Ours & \textbf{0.325} &   \textbf{0.605} &     \textbf{0.639} &     0.268 \\
Recon. loss & 0.068  & 0.211 & 0.138 & 0.211 \\
Ours w/o FFT       & 0.157 &   0.582 &     0.532 &     \textbf{0.284} \\
\bottomrule
\end{tabularx}
\caption{
\textbf{CelebA Ablations -- }
Our method provides significantly better performance than directly optimizing through the task-specific loss (the reconstruction loss).
Most evidently, directly learning to reconstruct results in low adherence to prompts, as shown by the low CLIP-T scores.
In fact, as shown in \cref{fig:recon_ablation}, the method starts ignoring the prompt.
Fast fine-tuning helps, but is not critical.
}
\label{tab:celeba_ablations}
\end{table}

\begin{table}[h!]
\centering
\begin{tabularx}{\linewidth}{@{}X c c c@{}}
\toprule
\textbf{Ablation} & CLIP-I & DINO & CLIP-T \\
\midrule
Ours  &   \textbf{0.664} &     \textbf{0.807} & 0.277 \\
Recon. loss  & 0.607 & 0.070 & 0.201 \\
Ours w/o FFT &   0.495 &     0.746 &     \textbf{0.285} \\
\bottomrule
\end{tabularx}
\caption{
\textbf{AFHQ Ablations -- }
Similar to the CelebA ablations, our full model is shown to perform best. As with the CelebA experiments, the Reconstruction loss ablation initially learned to simply output a copy of the condition image before breaking completely.
}
\label{tab:afhq_ablations}
\end{table}

\begin{table}[h!]
\centering
\begin{tabularx}{\columnwidth}{@{}X c @{}}
\toprule
Method & Preference (\%) \\
\midrule
Ours & \textbf{43.5} \\
DreamBooth & 35.1 \\
Textual Inversion & 21.4 \\
\bottomrule
\end{tabularx}
\caption{
User study results show that our method is preferred for identity preservation over benchmarks.
}

\label{fig:user_study}
\end{table}

\clearpage

\end{document}